\documentclass{article}
\usepackage{eaclap,avm}

\makeatletter
\def\maketitle{\par
 \begingroup
   \def\thefootnote{\fnsymbol{footnote}}
   \if@twocolumn
     \twocolumn[\@maketitle]
     \else \newpage
     \global\@topnum\z@
     \@maketitle \fi\thispagestyle{empty}\@thanks
 \endgroup
 \setcounter{footnote}{0}
 \let\maketitle\relax
 \let\@maketitle\relax
 \gdef\@thanks{}\gdef\@author{}\gdef\@title{}\let\thanks\relax}
\makeatother

\font\pr=cmvtt10 at 10pt

\newenvironment{code1}[0]
{\begin{center}
}
{\end{center}\vspace{-5ex}
}
\newenvironment{code3}[0]
{\begin{center}
 \tt
}
{\end{center}\vspace{-5ex}
}

\newtheorem{example}{Example}
\newcommand{\typ}[1]{\mbox{\it #1}}

\def\|{\mbox{\(|\)}}

\def\typ#1{\mbox{{\it #1}}}

\newsavebox{\boxtmp}

\newcommand{\bvmfont}{\tiny\rm}

\newcommand{\tagfont}{\scriptsize}

\newlength{\typeindent}

\newcommand{\bvm}[1]{\renewcommand{\arraystretch}{1.2}\mbox{%
\(\left[\mbox{\bvmfont\begin{tabular}{@{}l@{\,\hspace{0.4em}}l@{}}#1%
\end{tabular}}\right]\)}\renewcommand{\arraystretch}{1}}

\newcommand{\idx}[1]{\mbox{\fbox{\tagfont#1}\hspace{0.1em}}}

\newcounter{myfignr}

\input{psfig-scale}

\avmsortfont{\scriptsize}

\newlength{\figwidth}

\newsavebox{\figbox}

\newsavebox{\boxA} \newsavebox{\boxB} \newsavebox{\boxC}
\newsavebox{\boxD} \newsavebox{\boxE} \newsavebox{\boxF}

\newlength{\lenA} \newlength{\lenB} \newlength{\lenC}
\newlength{\lenD} \newlength{\lenE} \newlength{\lenF}

\title{\vspace{-0.5in}Selective Magic HPSG Parsing} \author{Guido
  Minnen\thanks{The presented research was carried out at the
    University of T\"ubingen, Germany, as part of the
    Sonderforschungsbereich 340.}
  \\
  Cognitive and Computing Sciences, University of Sussex\\
  Falmer, Brighton BN1 9QH\\
  United Kingdom\\
  Guido.Minnen@cogs.susx.ac.uk\\
  www.cogs.susx.ac.uk/lab/nlp/minnen/minnen.html}
\begin{document}
\maketitle
\begin{abstract}
\vspace{-.2cm}
  We propose a parser for constraint-logic grammars implementing
  HPSG that combines the advantages of dynamic bottom-up and
  advanced top-down control.  The parser allows the user to apply
  magic compilation to specific constraints in a grammar which as
  a result can be processed dynamically in a bottom-up and
  goal-directed fashion.  State of the art top-down processing
  techniques are used to deal with the remaining constraints.  We
  discuss various aspects concerning the implementation of the
  parser as part of a grammar development system.
\end{abstract}

\vspace{-.5cm}
\marginpar{\vspace*{-13.2cm}\hspace*{3.54cm}\mbox{\normalsize\it In
    Proceedings of the 9th Conference of the EACL, Bergen, Norway,
    June 1999.}}
\section{Introduction}
\label{sec1}
\vspace{-.1cm} In case space requirements of dynamic parsing often
outweigh the benefit of not duplicating sub-computations. We propose a
parser that avoids this drawback through combining the advantages of
dynamic bottom-up and advanced top-down control.\footnote{A more
  detailed discussion of various aspects of the proposed parser can be
  found in~\cite{Minnen:98}.}  The underlying idea is to achieve
faster parsing by avoiding tabling on sub-computations which are not
expensive.  The so-called \textit{selective magic parser} allows the
user to apply magic compilation to specific constraints in a grammar
which as a result can be processed dynamically in a bottom-up and
goal-directed fashion.  State of the art top-down processing
techniques are used to deal with the remaining constraints.

Magic is a compilation technique originally developed for 
goal-directed bottom-up processing of logic programs.  See, among 
others, (Ramakrishnan et al.  1992).  As shown in~\cite{Minnen:96} 
magic is an interesting technique with respect to natural language 
processing as it incorporates filtering into the logic underlying the 
grammar and enables elegant control independent filtering 
improvements.  In this paper we investigate the selective application 
of magic to {\it typed feature grammars} a type of constraint-logic 
grammar based on Typed Feature Logic (${\cal TFL}$; G\"otz, 1995).  
Typed feature grammars can be used as the basis for implementations of 
Head-driven Phrase Structure Grammar (HPSG; Pollard and Sag, 1994) as 
discussed in (G\"otz and Meurers, 1997a) and (Meurers and Minnen, 
1997).  Typed feature grammar constraints that are inexpensive to 
resolve are dealt with using the top-down interpreter of the ConTroll 
grammar development system~\cite{Goetz:Meurers:97b} which uses an 
advanced search function, an advanced selection function and 
incorporates a coroutining mechanism which supports delayed 
interpretation.
\nocite{Ramakrishnan:Srivastava:Sudarshan:92}
\nocite{Goetz:Meurers:97a} 
\nocite{Meurers:Minnen:97}
\nocite{Goetz:95}
\nocite{Pollard:Sag:94}

The proposed parser is related to the so-called {\it Lemma Table}
deduction system~\cite{Johnson:Doerre:95} which allows the user to
specify whether top-down sub-computations are to be tabled. In contrast
to Johnson and D\"{o}rre's deduction system, though, the selective
magic parsing approach combines top-down and bottom-up control
strategies.  As such it resembles the parser of the grammar
development system Attribute Language Engine (ALE) of
\cite{Carpenter:Penn:94}.  Unlike the ALE parser, though, the
selective magic parser does not presuppose a phrase structure backbone
and is more flexible as to which sub-computations are tabled/filtered.
\vspace{-.1cm}
\section{Bottom-up Interpretation of Magic-compiled Typed Feature Grammars}
\label{sec2}
We describe typed feature grammars and discuss their use in
implementing HPSG grammars.  Subsequently we present magic
compilation of typed feature grammars on the basis of an example
and introduce a dynamic bottom-up interpreter that can be used
for goal-directed interpretation of magic-compiled typed feature
grammars.
\subsection{Typed Feature Grammars}
\label{sec2_1}
A typed feature grammar consists of a signature and a set of definite 
clauses over the constraint language of equations of ${\cal 
TFL}$~\cite{Goetz:95} terms~\cite{Hoehfeld:Smolka:88} which we will 
refer to as ${\cal TFL}$ definite clauses.  Equations over ${\cal 
TFL}$ terms can be solved using (graph) unification provided they are in 
normal form.  \cite{Goetz:94} describes a normal form for ${\cal TFL}$ 
terms, where typed feature structures are interpreted as satisfiable 
normal form ${\cal TFL}$ terms.\footnote{This view of typed feature 
structures differs from the perspective on typed feature structures as 
modeling partial information as in~\cite{Carpenter:92}.  Typed feature 
structures as normal form ${\cal TFL}$ terms are merely syntactic 
objects.} The signature consists of a type hierarchy and a set of 
appropriateness conditions.
\begin{example}
\begin{em}
  The signature specified in figure~\ref{sig1} and~\ref{sig2} and the
  ${\cal TFL}$ definite clauses in figure~\ref{dcs} constitute an
  example of a typed feature grammar.
\begin{figure*}[hbtp]
\begin{center}
\mbox{\psfig{figure=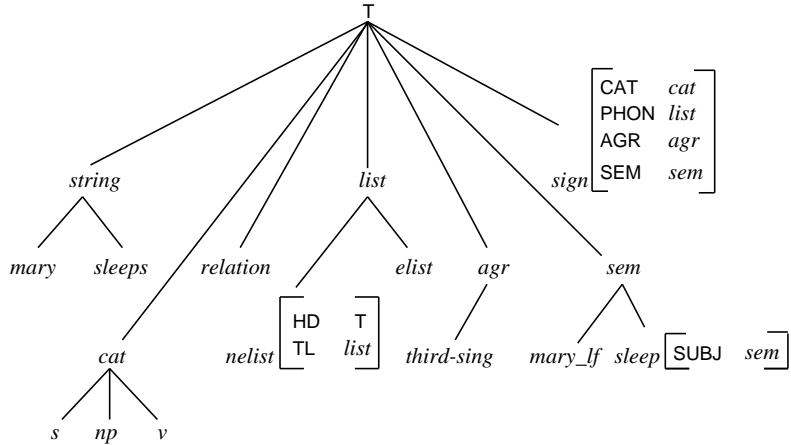,scale=75}}
\end{center}
\vspace{-.4cm}
\caption{\sl Example of a typed feature grammar signature (part 1)}
\label{sig1}
\end{figure*}
\begin{figure}[hbtp]
\begin{center}
\mbox{\psfig{figure=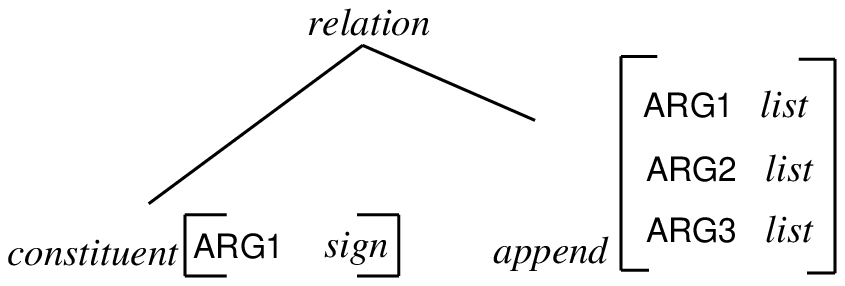,scale=75}}
\end{center}
\vspace{-.4cm}
\caption{\sl Example of a typed feature grammar signature (part 2)}
\label{sig2}
\end{figure}
\begin{figure}[hbtp]
\begin{small}
\begin{code1}
    \vspace*{-.5cm}  \parbox{0cm}{\begin{tabbing}
        \ \ \ \ \ \ \ \ \ \ \ \ \=\\
        (1) constituent({\bvm{{\tiny CAT} & \typ{s}\\PHON & \idx{1}\\SEM & \idx{5}}}):-\\[.1cm]
        \>constituent({\bvm{CAT & \typ{np}\\PHON & \idx{2}\\AGR &
            \idx{4}\\SEM & \idx{6}}}),\\[.1cm]
        \>constituent({\bvm{CAT & \typ{v}\\PHON & \idx{3}\\AGR & \idx{4}\\SEM & \idx{5}\bvm{SUBJ & \idx{6}}}}),\\[.1cm]
        \>append(\idx{2},\idx{3},\idx{1}).\\[.1cm]
        (2) constituent({\bvm{CAT & \typ{np}\\PHON & $\langle$ \typ{mary} $\rangle$\\AGR & \typ{third-sing}\\SEM & \typ{mary\_lf}}}).\\[.1cm]
        (3) constituent({\bvm{CAT & \typ{v}\\PHON & $\langle \typ{sleeps}\rangle$\\AGR & \typ{third-sing}\\SEM & \typ{sleep}}}).\\[.1cm]
        (4) append($\langle \rangle$, \idx{Ys}, \idx{Ys}).\\[.1cm]
        (5) append($\langle \idx{X} \mid \idx{Xs} \rangle$, \idx{Ys},
        $\langle \idx{X} \mid
        \idx{XsYs} \rangle$):-\\[.1cm]
        \>append(\idx{Xs},
        \idx{Ys}, \idx{XsYs}).\\
  \end{tabbing}}
\end{code1}
\end{small}
\caption{\sl Example of a set of ${\cal TFL}$ definite clauses}
\label{dcs}
\end{figure} 
We write ${\cal TFL}$ terms in normal form, i.~e., as typed feature
structures.  In addition, uninformative feature specifications are
ignored and typing is left implicit when immaterial to the example at
hand.  Equations between typed feature structures are removed by
simple substitution or tags indicating structure sharing.  Notice that
we also use non-numerical tags such as \idx{Xs} and \idx{XsYs}.  In
general all boxed items indicate structure sharing.  For expository
reasons we represent the ARG\textit{n} features of the {\pr append}
relation as separate arguments.
\end{em}
\end{example}

Typed feature grammars can be used as the basis for implementations of 
Head-driven Phrase Structure Grammar~\cite{Pollard:Sag:94}.%
\footnote{See~\cite{King:94b} for a discussion of the appropriateness 
of ${\cal TFL}$ for HPSG and a comparison with other feature logic 
approaches designed for HPSG.} \cite{Meurers:Minnen:97} propose a 
compilation of lexical rules into ${\cal TFL}$ definite clauses which 
are used to restrict lexical entries.  \cite{Goetz:Meurers:97b} 
describe a method for compiling implicational constraints into typed 
feature grammars and interleaving them with relational 
constraints.\footnote{\cite{Goetz:95} proves that this compilation 
method is sound in the general case and defines the large class of 
type constraints for which it is complete.} Because of space 
limitations we have to refrain from an example. The ConTroll grammar 
development system as described in~\cite{Goetz:Meurers:97b} implements 
the above mentioned techniques for compiling an HPSG theory into typed 
feature grammars.

\subsection{Magic Compilation}
\label{sec2_2}
Magic is a compilation technique for goal-directed bottom-up
processing of logic programs.  See, among others, (Ramakrishnan
et al.  1992).  Because magic compilation does not refer to the
specific constraint language adopted, its application is not
limited to logic programs/grammars: It can be applied to
relational extensions of other constraint languages such as typed
feature grammars without further adaptions.

Due to space limitations we discuss magic compilation by example
only.  The interested reader is referred
to~\cite{Nilsson:Maluszynski:95} for an introduction.
\begin{example}
\begin{em}
We illustrate magic compilation of typed feature grammars with respect 
to definite clause 1 in figure~\ref{dcs}. Consider the ${\cal TFL}$ 
definite clause in figure~\ref{magic-dcs}.
\begin{figure}[hbtp]
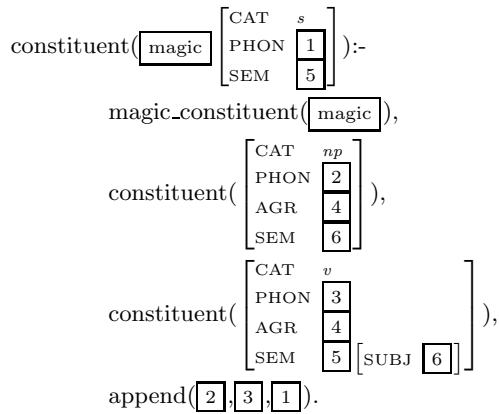

\begin{small}
\begin{code1}
    \vspace*{-.5cm}  \parbox{0cm}{\begin{tabbing}
        \ \ \ \ \ \ \ \ \ \ \ \ \=\\
        constituent(\idx{magic}{\bvm{CAT & \typ{s}\\PHON & \idx{1}\\SEM & \idx{5}}}):-\\[.1cm]
        \>magic\_constituent(\idx{magic}),\\[.1cm]
        \>constituent({\bvm{CAT & \typ{np}\\PHON & \idx{2}\\AGR &
            \idx{4}\\SEM & \idx{6}}}),\\[.1cm]
        \>constituent({\bvm{CAT & \typ{v}\\PHON & \idx{3}\\AGR & \idx{4}\\SEM & \idx{5}\bvm{SUBJ & \idx{6}}}}),\\[.1cm]
        \>append(\idx{2},\idx{3},\idx{1}).\\
  \end{tabbing}}
\end{code1}
\end{small}
\caption{\sl Magic variant of definite clause 1 in figure~\ref{dcs} }
\label{magic-dcs}
\end{figure}
As a result of magic compilation a magic literal is added to the 
right-hand side of the original definite clause.  Intuitively 
understood, this magic literal ``guards'' the application of the 
definite clause.  The clause is applied only when there exists a fact 
that unifies with this magic literal.\footnote{A fact can be a unit 
clause, i.~e., a ${\cal TFL}$ definite clause without right-hand side 
literals, from the grammar or derived using the rules in the grammar.  
In the latter case one also speaks of a passive edge.} The 
resulting definite clause is also referred to as the \textit{magic 
variant} of the original definite clause.

The definite clause in figure~\ref{seed} is the so-called 
\textit{seed} which is used to make the bindings as provided by the 
initial goal available for bottom-up processing.  In this case the 
seed corresponds to the initial goal of parsing the string `mary 
sleeps'.
\begin{figure}[hbtp]
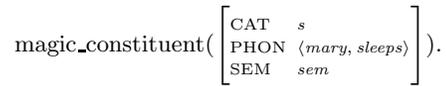

\begin{small}
\begin{code1}
    \vspace*{-.5cm}  \parbox{0cm}{\begin{tabbing}
        \ \ \ \ \ \ \ \ \ \ \ \ \=\\
        magic\_constituent({\bvm{CAT & \typ{s}\\PHON & $\langle \typ{mary}, \typ{sleeps} \rangle$\\SEM & \typ{sem}}}).\\
  \end{tabbing}}
\end{code1}
\end{small}
\caption{\sl Seed corresponding to the initial goal of 
parsing the string `mary sleeps'}
\label{seed}
\end{figure}
Intuitively understood, the seed makes available the bindings of the 
initial goal to the magic variants of the definite clauses defining a 
particular initial goal; in this case the magic variant of the 
definite clause defining a constituent of category `s'.  Only when 
their magic literal unifies with the seed are these clauses 
applied.\footnote{The creation of the seed can be postponed until run 
time, such that the grammar does not need to be compiled for every 
possible initial goal.}

The so-called \textit{magic rules} in figure~\ref{magic-rules} are 
derived in order to be able to use the bindings provided by the seed 
to derive new facts that provide the bindings which allow for a 
goal-directed application of the definite clauses in the grammar not 
directly defining the initial goal.
\begin{figure}[hbtp]
\begin{small}
\begin{code1}
    \vspace*{-.5cm}  \parbox{0cm}{\begin{tabbing}
        \ \ \ \ \ \ \ \ \ \ \ \ \=\\
        (1) magic\_constituent({\bvm{CAT & \typ{np}\\PHON & \typ{list}\\AGR &
            \typ{agr}\\SEM & \typ{sem}}}):-\\[.1cm]
        \>magic\_constituent({\bvm{CAT & \typ{s}\\PHON & \typ{list}\\SEM & \typ{sem}}}).\\[.1cm]
        (2) magic\_constituent({\bvm{CAT & \typ{v}\\PHON & \typ{list}\\AGR & \idx{4}\\SEM & \idx{5}\bvm{SUBJ & \idx{6}}}}):-\\[.1cm]
        \>magic\_constituent({\bvm{CAT & \typ{s}\\PHON & \typ{list}\\SEM & \idx{5}}}),\\[.1cm]
        \>constituent({\bvm{CAT & \typ{np}\\PHON & \typ{list}\\AGR &
            \idx{4}\\SEM & \idx{6}}}),\\[.1cm]
        (3) magic\_append(\idx{2},\idx{3},\idx{1}):-\\[.1cm]
        \>magic\_constituent({\bvm{CAT & \typ{s}\\PHON & \idx{1}\\SEM & \idx{5}}}),\\[.1cm]
        \>constituent({\bvm{CAT & \typ{np}\\PHON & \idx{2}\\AGR &
            \idx{4}\\SEM & \idx{6}}}),\\[.1cm]
        \>constituent({\bvm{CAT & \typ{v}\\PHON & \idx{3}\\AGR & \idx{4}\\SEM & \idx{5}\bvm{SUBJ & \idx{6}}}}).\\
  \end{tabbing}}
\end{code1}
\end{small}
\caption{\sl Magic rules resulting from applying magic 
  compilation to definite clause 1 in figure~\ref{dcs}}
\label{magic-rules}
\end{figure}
Definite clause 3, for example, can be used to derive a {\pr 
magic\_append} fact which percolates the relevant bindings of the 
seed/initial goal to restrict the application of the magic variant of 
definite clauses 4 and 5 in figure~\ref{dcs} (which are not 
displayed).
\end{em}
\end{example}

\subsection{Semi-naive Bottom-up Interpretation}
\label{sec2_3}
Magic-compiled logic programs/grammars can be interpreted in a
bottom-up fashion without losing any of the goal-directedness
normally associated with top-down interpretation using a
so-called \textit{semi-naive bottom-up} interpreter: A dynamic
interpreter that tables only complete intermediate results,
i.~e., facts or passive edges, and uses an agenda to avoid
redundant sub-computations.  The Prolog predicates in
figure~\ref{sbi} implement a semi-naive bottom-up
interpreter.\footnote{Definite clauses serving as data are
  encoded using the predicate {\tt definite\_clause}/1: {\tt
    definite\_clause((Lhs :- Rhs)).}, where {\tt Rhs} is a
  (possibly empty) list of literals.}  In this interpreter both
the table and the agenda are represented using
lists.\footnote{There are various other---more efficient---ways
  to implement a dynamic control strategy in Prolog.  See, for
  example, \cite{Shieber:Schabes:Pereira:95}.} The agenda keeps
track of the facts that have not yet been used to update the
table.  It is important to notice that in order to use the
interpreter for typed feature grammars it has to be adapted to
perform graph unification.\footnote{A term encoding of typed
  feature structures would enable the use of term unification
  instead. See, for example, \cite{Gerdemann:95c}. } We refrain
from making the necessary adaptions to the code for expository
reasons.

The table is initialized with the facts from the grammar.  Facts are
combined using a operation called {\it match}.  The match operation
unifies all but one of the right-hand side literals of a definite
clause in the grammar with facts in the table.  The remaining
right-hand side literal is unified with a newly derived fact, i.~e., a
fact from the agenda.  By doing this, repeated derivation of facts
from the same earlier derived facts is avoided.
\begin{figure}[htbp]
\begin{small}
\begin{code3}
  \parbox{0cm}{\begin{tabbing}
      semi\=\_naive\=\_interpret(Goal):-\\
      \>initialization(Agenda,Table0),\\
      \>update\_table(Agenda,Table0,Table),\\
      \>member(edge(Goal,[]),Table).\\[.1cm]
      update\_table([],Table,Table).\\
      update\_table([Edge|Agenda0],Table0,Table):-\\
      \>update\_table\_w\_edge(Edge,Edges,\\
      \>\>\hspace{3.2cm}Table0,Table1),\\
      \>append(Edges,Agenda0,Agenda),\\
      \>update\_table(Agenda,Table1,Table).\\[.1cm]
      update\_table\_w\_edge(Edge,Edges,Table0,Table):-\\
      \>findall( \=NewEdge,\\
      \>\>match(Edge,NewEdge,Table0),\\
      \>\>Edges),\\
      \>store(Edges,Table0,Table).\\[.1cm]
      sto\=re([],Table,Table):-\\
      store([Edge|Edges],Table0,Table):-\\
      \>member(GenEdge,Table0),\\
      \>{\scriptsize $\backslash +$} subsumes(GenEdge,Edge),\\
      \>store(Edges,[Edge|Table0],Table).\\
      store([\_|Edges],Table0,Table):-\\
      \>store(Edges,Table0,Table).\\[.1cm]
      init\=ialization(Edges,Edges):-\\
      \>findall( \=edge(Head,[]),\\
      \>\>definite\_clause((Head:- [])),\\
      \>\>Edges).\\[.1cm]
      comp\=letion(Edge,edge(Goal,[]),Table):-\\
      \>definite\_clause((Goal :- Body)),\\
      \>Edge = edge(F,[]),\\
      \>select(F,Body,R),\\
      \>edges(R,Table).\\[.1cm]
      edges([],\_).\\
      edges([Lit$|$Lits],Table):-\\
      \>member(edge(Lit,[]),Table),\\
      \>edges(Lits,Table).\\
\end{tabbing}}
\end{code3}
\end{small}
\caption{\sl Semi-naive bottom-up interpreter}
\label{sbi}
\end{figure}

\section{Selective Magic HPSG Parsing}
\label{sec3}
In case of large grammars the huge space requirements of dynamic
processing often nullify the benefit of tabling intermediate results.
By combining control strategies and allowing the user to specify how
to process particular constraints in the grammar the selective magic
parser avoids this problem.  This solution is based on the observation
that there are sub-computations that are relatively
cheap and as a result do not need tabling~\cite{Johnson:%
  Doerre:95,Vannoord:97}.

\subsection{Parse Type Specification}
Combining control strategies depends on a way to differentiate between 
types of constraints.  For example, the ALE parser~\cite{Carpenter:%
  Penn:94} presupposes a phrase structure backbone which can be used
to determine whether a constraint is to be interpreted bottom-up or
top-down.  In the case of selective magic parsing we use so-called
{\it parse types} which allow the user to specify how constraints in
the grammar are to be interpreted.  A literal (goal) is considered a
\textit{parse type literal (goal)} if it has as its single argument a
typed feature structure of a type specified as a parse
type.\footnote{The notion of a parse type literal is closely related
  to that of a \textit{memo literal} as in~\cite{Johnson:Doerre:95}.}

\begin{figure}[htbp]
 \begin{center}
\vspace{.2cm}
\hspace{-.3cm}
   \mbox{\psfig{figure=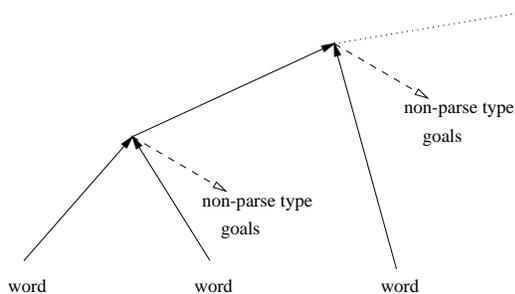,scale=65}}
   \label{schema-sel}
 \end{center}
\vspace{-.2cm}
\caption{\sl Schematic representation of the selective magic parsing 
 process}
\label{schema-parse}
\end{figure}

All types in the type hierarchy can be used as parse types.  This
way parse type specification supports a flexible filtering
component which allows us to experiment with the role of
filtering. However, in the remainder we will concentrate on a
specific class of parse types: We assume the specification of
type {\it sign} and its sub-types as parse types.\footnote{When a
  type is specified as a parse type, all its sub-types are
  considered as parse types as well.  This is necessary as
  otherwise there may exist magic variants of definite clauses
  defining a parse type goal for which no magic facts can be
  derived which means that the magic literal of these clauses can
  be interpreted neither top-down nor bottom-up.} This choice is
based on the observation that the constraints on type {\it sign}
and its sub-types play an important guiding role in the parsing
process and are best interpreted bottom-up given the lexical
orientation of HPSG.  The parsing process corresponding to such a
parse type specification is represented schematically in
figure~\ref{schema-parse}.
Starting from the lexical entries, i.~e., the ${\cal TFL}$
definite clauses that specify the word objects in the grammar,
phrases are built bottom-up by matching the parse type literals
of the definite clauses in the grammar against the edges in the
table.  The non-parse type literals are processed according to
the top-down control strategy described in section~\ref{advance}.

\subsection{Selective Magic Compilation}
\label{smc}
In order to process parse type goals according to a semi-naive magic
control strategy, we apply magic compilation selectively.  Only the
${\cal TFL}$ definite clauses in a typed feature grammar which define
parse type goals are subject to magic compilation.  The compilation
applied to these clauses is identical to the magic compilation
illustrated in section~\ref{sec2_1} except that we derive magic rules
only for the right-hand side literals in a clause which are of a parse
type.  The definite clauses in the grammar defining non-parse type
goals are not compiled as they will be processed using the top-down
interpreter described in the next section.

\subsection{Advanced Top-down Control}
\label{advance}
Non-parse type goals are interpreted using the standard
interpreter of the ConTroll grammar development
system~\cite{Goetz:Meurers:97b} as developed and implemented by
Thilo G\"otz. This advanced top-down interpreter uses a search
function that allows the user to specify the information on which
the definite clauses in the grammar are indexed.  An important
advantage of deep multiple indexing is that the linguist does not
have to take into account of processing criteria with respect to
the organization of her/his data as is the case with a standard
Prolog search function which indexes on the functor of the first
argument.

Another important feature of the top-down interpreter is its
use of a selection function that interprets deterministic goals,
i.~e., goals which unify with the left-hand side literal of
exactly one definite clause in the grammar, prior to
non-deterministic goals.  This is often referred to as
incorporating {\it deterministic closure}~\cite{Doerre:93}.
Deterministic closure accomplishes a reduction of the number of
choice points that need to be set during processing to a minimum.
Furthermore, it leads to earlier failure detection.

Finally, the used top-down interpreter implements a powerful
coroutining mechanism:\footnote{Coroutining appears under many
  different guises, like for example, {\it suspension}, {\it
    residuation}, {\it (goal)
freezing}, and {\it blocking}.  See also~\cite{Colmerauer:82,%
Naish:86}.} At run time the processing of a goal is postponed in case
it is insufficiently instantiated.  Whether or not a goal is
sufficiently instantiated is determined on the basis of so-called {\it
  delay patterns}.\footnote{In the literature delay patterns are
  sometimes also referred to as {\it wait declarations} or {\it block
    statements}.} These are specifications provided by the user that
indicate which restricting information has to be available before a
goal is processed.

\subsection{Adapted Semi-naive Bottom-up Interpretation} 
The definite clauses resulting from selective magic
transformation are interpreted using a semi-naive bottom-up
interpreter that is adapted in two respects. It ensures that
non-parse type goals are interpreted using the advanced top-down
interpreter, and it allows non-parse type goals that remain
delayed locally to be passed in and out of sub-computations in a
similar fashion as proposed by~\cite{Johnson:Doerre:95}.  In
order to accommodate these changes the adapted semi-naive
interpreter enables the use of edges which specify delayed goals.

Figure~\ref{adapcompletion} illustrates the adapted match
operation. %
\begin{figure}[htbp]
\begin{small}
\begin{code3}
      \parbox{0cm}{\begin{tabbing}
              mat\=ch(Edge,edge(Goal,Delayed),Table):-\\
              \>definite\_clause((Goal :- Body)),\\
              \>select(Lit,Body,Lits),\\
              \>parse\_type(Lit),\\
              \>Edge = edge(Lit,Delayed0),\\
              \>edges(Lit,Table,Delayed0,TopDown),\\
              \>advanced\_td\_interpret(TopDown,Delayed).\\
              match(Edge,edge(Goal,Delayed),Table):-\\
              \>definite\_clause((Goal :- TopDown)),\\
              \>advanced\_td\_interpret(TopDown,Delayed).\\
\end{tabbing}}
\end{code3}
\end{small}
\caption{\sl Adapted definition of {\tt match}/3}
\label{adapcompletion}
\end{figure}%
\begin{figure*}[htb]
\begin{center}
\mbox{\psfig{figure=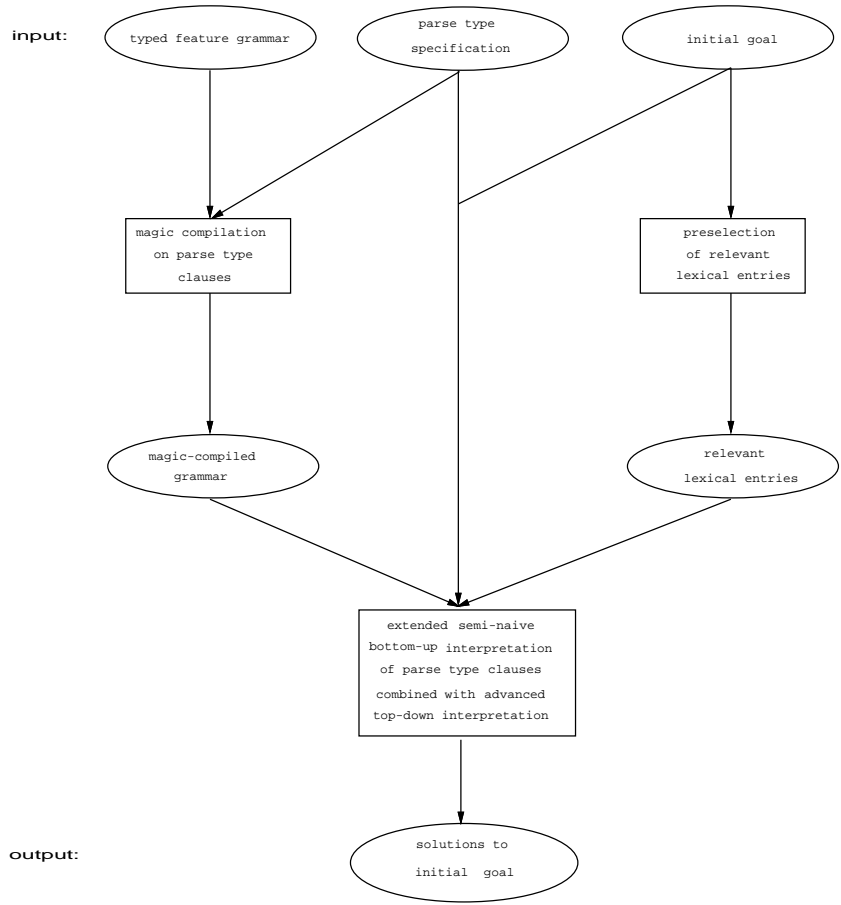,scale=60}}
\end{center}
\caption{\sl Setup of the ConTroll magic component}
\label{setup1}
\end{figure*}%
The first defining clause of {\tt match}/3 passes delayed and
non-parse type goals of the definite clause under consideration
to the advanced top-down interpreter via the call to {\tt
  advanced\_td\_interpret}/2 as the list of goals {\tt
  TopDown}.\footnote{The definition of {\tt match}/3 assumes that
  there exists a strict ordering of the right-hand side literals
  in the definite clauses in the grammar, i.~e., parse type
  literals always preced e non-parse type literals.}  The second
defining clause of {\tt match}/3 is added to ensure all
right-hand side literals are directly passed to the advanced
top-down interpreter if none of them are of a parse type.

Allowing edges which specify delayed goals necessitates the
adaption of the definition of {\tt edges}/3. When a parse type
literal is matched against an edge in the table, the delayed
goals specified by that edge need to be passed to the top-down
interpreter.  Consider the definition of the predicate {\tt
  edges} in figure~\ref{adapedges}.
\begin{figure}[htbp]
\begin{small}
\begin{code3}
      \parbox{0cm}{\begin{tabbing}
              edg\=es([Lit$|$Lits],Table,Delayed0,TopDown):-\\
              \>parse\_type(Lit),\\
              \>member(edge(Lit,Delayed1),Table),\\
              \>append(Delayed0,Delayed1,Delayed).\\
              \>edges(Lit,Table,Delayed,TopDown).\\
              edges([],\_,Delayed,TopDown):-\\
              \>append(Delayed,Lit,TopDown).\\
\end{tabbing}}
\end{code3}
\end{small}
\caption{\sl Adapted definition of {\tt edges}/4}
\label{adapedges}
\end{figure}
The third argument of the definition of {\tt edges}/4 is used to 
collect delayed goals.  When there are no more parse type literals in 
the right-hand side of the definite clause under consideration, the 
second defining clause of {\tt edges}/4 appends the collected delayed 
goals to the remaining non-parse type literals.  Subsequently, the 
resulting list of literals is passed up again for advanced top-down 
interpretation.

\section{Implementation}
\label{sec4}
The described parser was implemented as part of the ConTroll grammar
development system~\cite{Goetz:Meurers:97b}.  Figure~\ref{setup1}
shows the overall setup of the ConTroll magic component. The Controll
magic component presupposes a parse type specification and a set of
delay patterns to determine when non-parse type constraints are to be
interpreted. At run-time the goal-directedness of the selective magic
parser is further increased by means of using the phonology of the
natural language expression to be parsed as specified by the initial
goal to restrict the number of facts that are added to the table
during initialization.  Only those facts in the grammar corresponding
to lexical entries that have a value for their phonology feature that
appears as part of the input string are used to initialize the table.

The ConTroll magic component was tested with a larger ($>$ 5000 lines)
HPSG grammar of a sizeable fragment of German.  This grammar provides
an analysis for simple and complex verb-second, verb-first and
verb-last sentences with scrambling in the mittelfeld, extraposition
phenomena, wh-movement and topicalization, integrated verb-first
parentheticals, and an interface to an illocution theory, as well as
the three kinds of infinitive constructions, nominal phrases, and
adverbials~\cite{Hinrichs:Meurers:Richter:Sailer:Winhart:97}.

As the test grammar combines sub-strings in a non-concatenative fashion,
a preprocessor is used that chunks the input string into linearization
domains. This way the standard ConTroll interpreter (as described in
section~\ref{advance}) achieves parsing times of around 1-5 seconds
for 5 word sentences and 10--60 seconds for 12 word
sentences.\footnote{Parsing with such a grammar is difficult in any
  system as it does neither have nor allow the extraction of a phrase
  structure backbone.} The use of magic compilation on all grammar
constraints, i.e., tabling of all sub-computations, leads to an
vast increase of parsing times.  The selective magic HPSG parser,
however, exhibits a significant speedup in many cases. For example,
parsing with the module of the grammar implementing the analysis of
nominal phrases is up to nine times faster.  At the same time though
selective magic HPSG parsing is sometimes significantly slower.  For
example, parsing of particular sentences exhibiting adverbial
subordinate clauses and long extraction is sometimes more than nine
times slower.  We conjecture that these ambiguous results are due to
the use of coroutining: As the test grammar was implemented using the
standard ConTroll interpreter, the delay patterns used presuppose a
data-flow corresponding to advanced top-down control and are not
fine-tuned with respect to the data-flow corresponding to the
selective magic parser.

Coroutining is a flexible and powerful facility used in many grammar
development systems and it will probably remain indispensable in
dealing with many control problems despite its various
disadvantages.\footnote{Coroutining has a significant run-time
  overhead caused by the necessity to check the instantiation status
  of a literal/goal.  In addition, it demands the procedural
  annotation of an otherwise declarative grammar.  Finally,
  coroutining presupposes that a grammar writer possesses substantial
  processing expertise.} The test results discussed above indicate
that the comparison of parsing strategies can be seriously hampered by
fine-tuning parsing using delay patterns.  We believe therefore that
further research into the systematics underlying coroutining would be
desirable.

\section{Concluding Remarks}
\label{sec5}
We described a selective magic parser for typed feature grammars
implementing HPSG that combines the advantages of dynamic bottom-up
and advanced top-down control.  As a result the parser avoids the
efficiency problems resulting from the huge space requirements of
storing intermediate results in parsing with large grammars. The
parser allows the user to apply magic compilation to specific
constraints in a grammar which as a result can be processed
dynamically in a bottom-up and goal-directed fashion.  State of the
art top-down processing techniques are used to deal with the remaining
constraints.  We discussed various aspects concerning the
implementation of the parser which was developed as part of the
grammar development system ConTroll.

\section*{Acknowledgments}
The author gratefully acknowledges the support of the SFB 340 project
B4 ``From Constraints to Rules: Efficient Compilation of HPSG'' funded
by the German Science Foundation, and the project ``PSET: Practical
Simplification of English Text", a three-year project funded by the UK
Engineering and Physical Sciences Research Council (GR/L53175), and
Apple Computer Inc.. The author wishes to thank Dale Gerdemann and
Erhard Hinrichs and the anonymous reviewers for comments and
discussion.  Of course, the author is responsible for all remaining
errors.


\begin{thebibliography}{}
\bibitem[\protect\citename{Carpenter and Penn}1994]{Carpenter:Penn:94}
Bob Carpenter and Gerald Penn.
\newblock 1994.
\newblock ALE -- The Attribute Logic Engine, User's guide, version 2.0.2.
\newblock Technical report, Carnegie Mellon University, Pittsburgh, Pennsylvania, USA.
\vspace{-.1cm}

\bibitem[\protect\citename{Carpenter}1992]{Carpenter:92}
Bob Carpenter.
\newblock 1992.
\newblock {\em The Logic of Typed Feature Structures - With Applications to
  Unification Grammars, Logic Programs and Constraint Resolution}.
\newblock Cambridge University Press, New York, USA.
\vspace{-.1cm}

\bibitem[\protect\citename{Colmerauer}1982]{Colmerauer:82}
Alain Colmerauer.
\newblock 1982.
\newblock PrologII: Manuel de r\'{e}f\'{e}rence et mod\`{e}le th\'{e}orique.
\newblock Technical report, Groupe d'Intelligence Artificielle, Facult\'{e} de
  Sciences de Luminy, Marseille, France.
\vspace{-.1cm}

\bibitem[\protect\citename{D\"orre}1993]{Doerre:93}
Jochen D\"orre.
\newblock 1993.
\newblock Generalizing Earley Deduction for Constraint-based Grammars.
\newblock In Jochen D\"orre and Michael Dorna (eds.), 1993.  {\em Computational 
Aspects of Constraint-Based Linguistic Description I}. DYANA-2, Deliverable R1.2.A.
\vspace{-.1cm}

\bibitem[\protect\citename{Gerdemann}1995]{Gerdemann:95c}
Dale Gerdemann.
\newblock 1995.
\newblock Term Encoding of Typed Feature Structures.
\newblock In {\em Proceedings of the Fourth International Workshop on Parsing
  Technologies}, Prague, Czech Republic.
\vspace{-.1cm}

\bibitem[\protect\citename{G\"otz and Meurers}1997a]{Goetz:Meurers:97a}
Thilo G\"otz and Detmar Meurers.
\newblock 1997a.
\newblock Interleaving Universal Principles and Relational Constraints over
  Typed Feature Logic.
\newblock In {\em ACL/EACL Proceedings}, Madrid, Spain.
\vspace{-.1cm}

\bibitem[\protect\citename{G\"otz and Meurers}1997b]{Goetz:Meurers:97b}
Thilo G\"otz and Detmar Meurers.
\newblock 1997b.
\newblock The ConTroll System as Large Grammar Development Platform.
\newblock In {\em Proceedings of the ACL Workshop on Computational Environments for Grammar
  Development and Linguistic Engineering}, Madrid, Spain.
\vspace{-.1cm}

\bibitem[\protect\citename{G\"otz}1994]{Goetz:94}
Thilo G\"otz.
\newblock 1994.
\newblock A Normal Form for Typed Feature Structures.
\newblock Technical report SFB 340 nr.~40, University of T\"ubingen, Germany.
\vspace{-.1cm}

\bibitem[\protect\citename{G\"otz}1995]{Goetz:95}
Thilo G\"otz.
\newblock 1995.
\newblock Compiling HPSG Constraint Grammars into Logic Programs.
\newblock In {\em Proceedings of the Workshop on Computational Logic for Natural Language Processing}, Edinburgh, UK.%
\vspace{-.1cm}%
\bibitem[\protect\citename{Hinrichs \bgroup et al.\egroup
  }1997]{Hinrichs:Meurers:Richter:Sailer:Winhart:97}
Erhard Hinrichs, Detmar Meurers, Frank Richter, Manfred Sailer, and Heike
  Winhart.
\newblock 1997.
\newblock Ein HPSG-fragment des Deutschen, Teil 1: Theorie.
\newblock Technical report SFB 340~95, University of T\"ubingen, Germany.
\vspace{-.1cm}

\bibitem[\protect\citename{H\"ohfeld and Smolka}1988]{Hoehfeld:Smolka:88}
Markus H\"ohfeld and Gert Smolka.
\newblock 1988.
\newblock Definite Relations over Constraint Languages.
\newblock Technical Report~53, IBM, Germany.
\vspace{-.1cm}

\bibitem[\protect\citename{Johnson and D\"orre}1995]{Johnson:Doerre:95}
Mark Johnson and Jochen D\"orre.
\newblock 1995.
\newblock Memoization of Coroutined Constraints.
\newblock In {\em ACL Proceedings}, Cambridge, Massachusetts, USA.
\vspace{-.1cm}

\bibitem[\protect\citename{King}1994]{King:94b}
Paul King.
\newblock 1994.
\newblock Typed Feature Structures as Descriptions.
\newblock In {\em Proceedings of of the 15th Conference on Computational
  Linguistics}, Kyoto, Japan.
\vspace{-.1cm}

\bibitem[\protect\citename{Meurers and Minnen}1997]{Meurers:Minnen:97}
Detmar Meurers and Guido Minnen.
\newblock 1997.
\newblock A Computational Treatment of Lexical Rules in HPSG as Covariation in
  Lexical Entries.
\newblock {\em Computational Linguistics}, 23(4).
\vspace{-.1cm}

\bibitem[\protect\citename{Minnen}1996]{Minnen:96}
Guido Minnen.%
\newblock 1996.%
\newblock Magic for Filter Optimization in Dynamic Bottom-up Processing.%
\newblock In {\em ACL Proceedings}, Santa Cruz, California, USA.%
\vspace{-.1cm}%
%
\bibitem[\protect\citename{Minnen}1998]{Minnen:98}
Guido Minnen.
\newblock 1998.
\newblock {\em Off-line Compilation for Efficient Processing with
  Constraint-logic Grammars}.
\newblock {Ph.D.} thesis, University of T\"ubingen, Germany.
\newblock Technical report SFB 340 nr.~130.%
\vspace{-.1cm}

\bibitem[\protect\citename{Naish}1986]{Naish:86}
Lee Naish.
\newblock 1986.
\newblock {\em Negation and Control in Prolog}.
\newblock Springer-Verlag, Berlin, Germany.
\vspace{-.1cm}

\bibitem[\protect\citename{Nilsson and
  Ma{\l}uszynski}1995]{Nilsson:Maluszynski:95}
Ulf Nilsson and Jan Ma{\l}uszynski.
\newblock 1995.
\newblock {\em Logic, Programming and Prolog}.
\newblock John Wiley \& Sons, Chichester, UK, 2nd edition.
\vspace{-.1cm}

\bibitem[\protect\citename{Pollard and Sag}1994]{Pollard:Sag:94}
Carl Pollard and Ivan Sag.
\newblock 1994.
\newblock {\em Head-Driven Phrase Structure Grammar}.
\newblock University of Chicago Press, Chicago, Illinois, USA.
\vspace{-.1cm}

\bibitem[\protect\citename{Ramakrishnan \bgroup et al.\egroup
  }1992]{Ramakrishnan:Srivastava:Sudarshan:92}
Raghu Ramakrishnan, Divesh Srivastava, and S.~Sudarshan.
\newblock 1992.
\newblock Efficient Bottom-up Evaluation of Logic Programs.
\newblock In Joos Vandewalle (ed.), 1992. {\em The State of 
the Art in Computer Systems and Software Engineering}. Kluwer 
Academic Publishers.
\vspace{-.1cm}

\bibitem[\protect\citename{Shieber \bgroup et al.\egroup
  }1995]{Shieber:Schabes:Pereira:95}
Stuart Shieber, Yves Schabes, and Fernando Pereira.
\newblock 1995.
\newblock Principles and Implementation of Deductive Parsing.
\newblock {\em Journal of Logic Programming}, 24(1-2).
\vspace{-.1cm}

\bibitem[\protect\citename{van Noord}1997]{Vannoord:97}
Gertjan van Noord.
\newblock 1997.
\newblock An Efficient Implementation of the Head-corner Parser.
\newblock {\em Computational Linguistics}, 23(3).

\end{thebibliography}
\end{document}